\documentclass[10pt,twocolumn,letterpaper]{article}

\usepackage{cvpr}
\usepackage{times}
\usepackage{epsfig}
\usepackage{graphicx}
\usepackage{amsmath}
\usepackage{amssymb}
\usepackage{booktabs}

\usepackage[pagebackref=true,breaklinks=true,letterpaper=true,colorlinks,bookmarks=false]{hyperref}

\cvprfinalcopy 

\def\cvprPaperID{2} 

\ifcvprfinal\fi
\begin{document}

\title{Early Detection of Injuries in MLB Pitchers from Video}

\author{AJ Piergiovanni and Michael S. Ryoo\\
Department of Computer Science, Indiana University, Bloomington, IN 47408 \\
\texttt{\{ajpiergi,mryoo\}@indiana.edu}\\
}

\maketitle

\begin{abstract}
   Injuries are a major cost in sports. Teams spend millions of dollars every year on players who are hurt and unable to play, resulting in lost games, decreased fan interest and additional wages for replacement players. Modern convolutional neural networks have been successfully applied to many video recognition tasks. In this paper, we introduce the problem of injury detection/prediction in MLB pitchers and experimentally evaluate the ability of such convolutional models to detect and predict injuries in pitches only from video data. We conduct experiments on a large dataset of TV broadcast MLB videos of 20 different pitchers who were injured during the 2017 season. We experimentally evaluate the model's performance on each individual pitcher, how well it generalizes to new pitchers, how it performs for various injuries, and how early it can predict or detect an injury. 
\end{abstract}

\section{Introduction}

Injuries in sports is a major cost. When a start player is hurt, not only does the team continue paying the player, but also impacts the teams performance and fan interest. In the MLB, teams spend an average of \$450 million on players on the disabled list and an additional \$50 million for replacement players each year, an annual average \$500 million per year \cite{conte2016injury}. In baseball, pitcher injuries are some of the most costly and common, estimated as high as \$250 million per-year \cite{forbesinj}, about half the total cost of injuries in the MLB.

As a result, there are many studies on the causes, effects and recovery times of injuries caused by pitching. Mehdi et al. \cite{mehdi2016latissimus} studied the duration of lat injuries (back muscle and tendon area) in pitchers, finding an average recovery time of 100 days without surgery and 140 days for pitchers who needed surgery. Marshall et al.  \cite{marshall2018implications} found pitchers with core injuries took an average of 47 days to recover and 37 days for hip/groin injuries. These injuries not only effect the pitcher, but can also result in the team losing games and revenue. Pitching is a repetitive action; starting pitchers throw roughly 2500 pitches per-season in games alone - far more when including warm-ups, practice, and spring training. Due to such high use, injuries in pitchers are often caused by overuse \cite{lyman2002effect} and early detection of injuries could reduce severity and recovery time \cite{hreljac2005etiology,hergenroeder1998prevention}.

Modern computer vision models, such as convolutional neural networks (CNNs), allow machines to make intelligent decisions directly from visual data. 
Training a CNN to accurately detect injuries in pitches from purely video data would be extremely beneficial to teams and athletes, as they require no sensors, tests, or monitoring equipment other than a camera. A CNN trained on videos of pitchers would be able to detect slight changes in their form that could be a early sign of an injury or even cause an injury. The use of computer vision to monitor athletes can provide team physicians, trainers and coaches additional data to monitor and protect athletes.

CNN models have already been successfully applied to many video recognition tasks, such as activity recognition \cite{carreira2017quo}, activity detection \cite{piergiovanni2018super}, and recognition of activities in baseball videos \cite{mlbyoutube2018}. In this paper, we introduce the problem of injury detection/prediction in MLB pitchers and experimentally evaluate the ability of CNN models to detect and predict injuries in pitches from only video data.

\section{Related Work}
\paragraph{Video/Activity Recognition}
Video activity recognition is a popular research topic in computer vision~\cite{aggarwal11,karpathy2014large,simonyan2014two,wang2011action,ryoo13}. Early works focused on hand-crafted features, such as dense trajectories~\cite{wang2011action} and showed promising results. Recently, convolutional neural networks (CNNs) have out-performed the hand-crafted approaches \cite{carreira2017quo}. A standard multi-stream CNN approaches takes input of RGB frames and optical flows \cite{simonyan2014two,repflow2019} or RGB frames at different frame-rates \cite{feichtenhofer2018slowfast} which are used for classification, capturing different features. 3D (spatio-temproal) convolutional models have been trained for activity recognition tasks~\cite{tran2014c3d,carreira2017quo,piergiovanni2018evolving}. To train these CNN models, large scale datasets such as Kinetics~\cite{kay2017kinetics} and Moments-in-Time \cite{monfort2019moments} have been created.

Other works have explored using CNNs for temporal activity detection \cite{piergiovanni2018evolving, shou2017cdc, dave2017predictive} and studied the use of temporal structure or pooling for recognition \cite{piergiovanni2017learning, ng2015beyond}. Current CNNs are able to perform quite well on a variety of video-based recognition tasks.

\paragraph{Injury detection and prediction}
Many works have studied prediction and prevention of injuries in athletes by developing models based on simple data (e.g., physical stats or social environment) \cite{andersen1988model,ivarsson2017psychosocial} or cognitive and psychological factors (stress, life support, identity, etc.) \cite{maddison2005psychological,falkstein1999prediction}. Others made predictions based on measured strength before a season \cite{pontillo2014prediction}. Placing sensors on players to monitor their movements has been used to detect pitching events, but not injury detection or prediction \cite{lapinski2009distributed,murray2017automatic}. Further, sonography (ultra-sound) of elbows has been used to detect injuries by human experts \cite{harada2006using}.

To the best of our knowledge, there is no work exploring real-time injury detection in game environments. Further, our approach requires no sensors other than a camera. Our model makes predictions from only the video data.

\section{Data Collection}

Modern CNN models require sufficient amount of data (i.e., samples) for both their training and evaluation.
As pitcher injuries are fairly rare, especially compared to the number of pitches thrown while not injured, the collection and preparation of data is extremely important.
There is a necessity to best take advantage of such example videos while removing non-pitch related bias in the data.

In this work, we consider the task of injury prediction as a binary classification problem. That is, we label a video clip of a pitch either as `healthy' or `injured'. We assume the last $k$ pitches thrown before a pitcher was placed on the disabled list to be `injured' pitches.
If an injury occurred during practice or other non-game environment, we do not include that data in our dataset (as we do not have access to video data outside of games). We then collect videos of the TV broadcast of pitchers from several games not near the date of injury as well as the game they were injured in. This provides sufficient `healthy' as well as `injured' pitching video data.

\begin{figure}
    \centering
    \includegraphics[width=\linewidth]{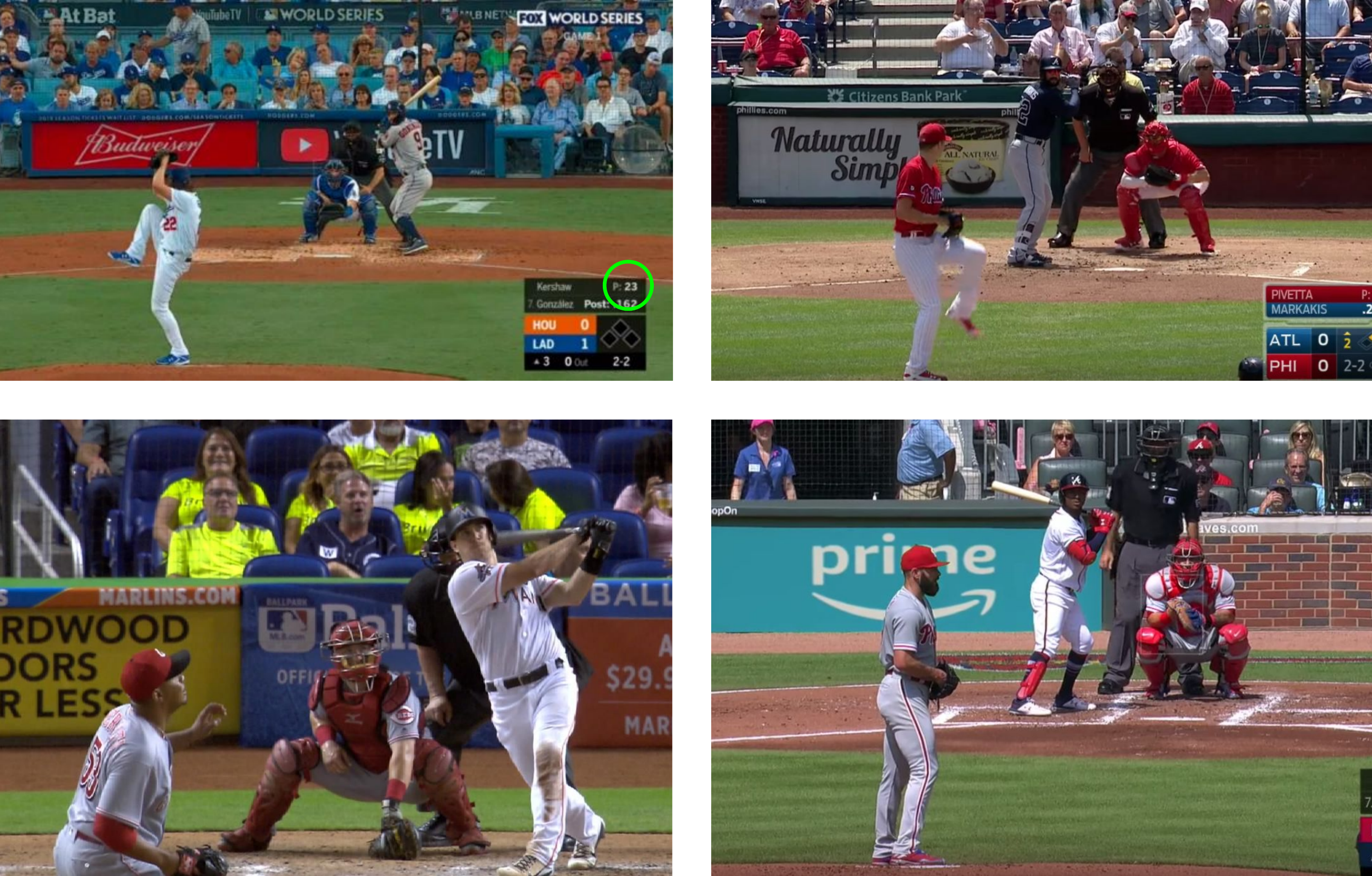}
    \caption{Examples of (top-left) the the pitch count being visible, (bottom-left) unique colored shirts in the background and (right) the same pitcher in different uniforms and ballparks. These features allow for the model to overfit to data that will not generalize or properly represent whether or not the pitcher is injured.}
    \label{fig:overfit}
\end{figure}
\begin{figure*}
    \centering
    \includegraphics[width=\linewidth]{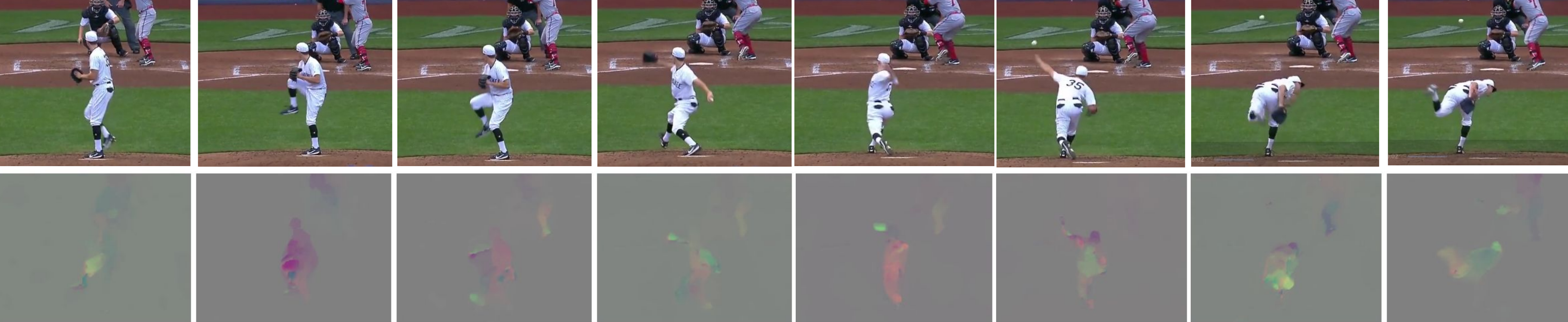}
    \caption{Example of cropped RGB frames and the computed optical flows.}
    \label{fig:flow}
\end{figure*}

The challenge in our dataset construction is that simply taking the broadcast videos and (temporally) segmenting each pitch interval is not sufficient. We found that the model often overfits to the pitch count on the scoreboard, the teams playing, or the exact pitcher location in the video (as camera position can slightly vary between ballparks and games), rather than focusing on the actual pitching motion. Spatially cropping the pitching region is also insufficient, as there could be an abundant amount of irrelevant information in the spatial bounding box. The model then overfits to the jersey of the pitcher, time of day or weather (based on brightness, shadows, etc.) or even a fan in a colorful shirt in the background (see Fig. \ref{fig:overfit} for examples). While superstitious fans may find these factors meaningful, they do not have any real impact on the pitcher or his injuries.

To address all these challenges, we first crop the videos to a bounding box containing just the pitcher. We then convert the images to greyscale and compute optical flow, as it capture high-resolution motion information while being invariant to appearance (i.e., jersey, time of day, etc.). Optical flow has commonly been used for activity detection tasks \cite{simonyan2014two} and is beneficial as it captures appearance invariant motion features \cite{sevilla2018integration}. We then use the optical flow frames as input to a CNN model trained for our binary injury classification task. This allows the model to predict a pitcher's injury based solely on motion information, ignoring the irrelevant features. Examples of the cropped frames and optical flows are shown in Fig. \ref{fig:flow}.

\paragraph{Dataset} Our dataset consists of pitches from broadcast videos for 30 games from the 2017 MLB season. It contains injuries from 20 different pitchers, 4 of which had multiple injuries in the same season. Each pitcher has an average of 100 healthy pitches from games not near where they were injured as well as pitches from the game they in which they were injured. The data contains 12 left-handed pitchers and 8 right-handed pitchers, 10 different injuries (back strain, arm strain, finger blister, shoulder strain, UCL tear, intercoastal strain, sternoclavicular joint, rotator cuff, hamstring strain, and groin strain). There are 5479 pitches in the dataset, about 273 per-pitcher, providing sufficient data to train a video CNN. When using $k=20$, resulting in 469 `injured' pitches and 5010 healthy pitches, as some pitchers threw less than 20 pitches before being injured.

\section{Approach}
We use a standard 3D spatio-temporal CNN trained on the optical flow frames. Specifically, we use I3D \cite{carreira2017quo} with 600 optical flow frames as input with resolution of $460\times 600$ (cropped to just the pitcher from $1920\times 1080$ video) from a 10 second clip of a pitch at 60 fps. We use high frame-rate and high-resolution inputs to allow the model to learn the very small differences between `healthy' and `injured' pitches. We initialize I3D with the optical flow stream pre-trained on the Kinetics dataset \cite{kay2017kinetics} to obtain good initial weights. 

We train the model to minimize the binary cross entropy:
\begin{equation}
    \mathcal{L} = \sum_i \left( y_i \log p_i + (1-y_i) \log (1-p_i) \right)
\end{equation}
where $y_i$ is the label (injured or not) and $p_i$ is the models prediction for sample $i$. We train for 100 epochs with a learning rate of 0.1 that is decayed by a factor of 10 every 25 epochs. We use dropout set at 0.5 during training.

\section{Experiments}
We conduct extensive experiments to determine what the model (i.e., I3D video CNN) is capable of learning and how well it generalizes. We compare (1) learning models per-pitcher and test how well they generalize to other pitchers, (2) models learned from a set of lefty or righty pitchers, and (3) models trained on a set of several pitchers. We evaluate on both seen and unseen pitchers and seen and unseen injuries. We also compare models trained on specific injury types (e.g. back strain, UCL injury, finger blisters, etc.) and analyze how early we can detect an injury solely from video data.

Since this is a binary classification task, as the evaluation metric, we report:

\begin{itemize}
    \item Accuracy (correct examples/total examples)
    \item Precision (correct injured/predicted injured)
    \item Recall (correct injured/total injured)
    \item $F_1$ scores
\end{itemize}
where
\begin{equation}
    F_1 = \frac{1}{0.5\left(\cfrac{1}{recall}+\cfrac{1}{precision}\right)}
\end{equation}
All values are measured between 0 and 1, where 1 is perfect for the given measure.

\subsection{Per-player model}
We first train a model for each pitcher in the dataset. We consider the last 20 pitches thrown by a pitcher before being placed on the disabled list (DL) as `injured' and all other pitches thrown by the pitcher as healthy. We use half the `healthy' pitches and half the `injured' pitches as training data and the other half as test data. All the pitchers in the test dataset were seen during training. In Table \ref{tab:per-pitcher} we compare the results of our model for 10 different pitchers. For some pitchers, such as Clayton Kershaw or Boone Logan, the model was able to accurately detect their injury, while for other pitchers, such as Aaron Nola, the model was unable to reliably detect the injury.

\begin{table}[]
    \centering
    \begin{tabular}{l|cccc}
        \toprule
        Pitcher & Acc & Prec & Rec & $F_1$  \\
        \midrule
        Aaron Nola & .92 & .50 & .34 & .42 \\
        Clayton Kershaw & .96 & 1. & .75 & .86 \\
        Corey Kluber & .95 & 1. & .33 & .50 \\
        Aroldis Chapman & .83 & .50 & 1. & .67\\
        Boone Logan & .98 & .96 & .97 & .97 \\
        Brett Anderson & .96 & .75 & .97 & .85 \\
        Brandon Finnegan & .95 & .75 & .99 & .87 \\
        Austin Brice & .87 & .89 & .75 & .82 \\
        AJ Griffin & .91 & .98 & .33 & .50\\
        Adalberto Mejia & .96 & .83 & .78 & .81\\
        \midrule
        Average & .93 & .82 & .72 & .75 \\
         \bottomrule
    \end{tabular}
    \caption{Results of predicting a pitcher's injury where the last 20 pitches thrown are injured. A model was trained for each pitcher. The model performs well for some pitchers and poorly for others.}
    \label{tab:per-pitcher}
\end{table}

To determine how well the models generalize, we evaluate the trained models on a different pitcher. Our results are shown in Table \ref{tab:pitcher-to-pitcher}. We find that for some pitchers, the transfer works reasonable well, such as Libertore and Wainwright or Brice and Wood. However, for other pitchers, it does not generalize at all. This is not surprising, as pitchers have various throwing motions, use different arms, etc., in fact, it is quite interesting that it generalizes at all. 

\begin{table}[]
    \centering
    \begin{tabular}{ll|cccc}
        \toprule
        Train Pitcher & Test Pitcher & Acc & Prec & Rec & $F_1$  \\
        \midrule
        Liberatore & Wainwright  & .33 & .17 & .87 & .27  \\
        Wainwright & Liberatore  & .29 & .23 & .85 & .24  \\
        Wood & Liberatore & .63 & 0. & 0. & 0. \\
        Chapman & Finnegan & .75 & .23 & .18 & .19 \\
        Brice & Wood & .43 & .17 & .50 & .24 \\
        Wood & Brice & .42 & .15 & .48 & .22 \\
        Chapman & Bailey & .23 & 0. & 0. & 0.\\
        Mejia & Kulber & .47 & 0. & 0. & 0.\\
         \bottomrule
    \end{tabular}
    \caption{Comparing how well a model trained on one pitcher transfers to another pitcher. As throwing motions vary greatly between pitchers, the models do not generalize too well.}
    \label{tab:pitcher-to-pitcher}
\end{table}

\subsection{By pitching arm}
To further examine how well the model generalizes, we train the model on the 12 left handed (or  8 right handed) pitchers, half the data is used for training and the other half  of held-put pitches is used for evaluation. This allows us to determine if the model is able to learn injuries and throwing motions for multiple pitchers or if it must be specific to each pitcher. Here, all the test data is of pitchers seen during training. We also train a model on all 20 pitchers and test on held-out pitches. Our results are shown in Table \ref{tab:by-arm1}. We find that these models perform similarly to the per-pitcher model, suggesting that the model is able to learn multiple pitchers' motions. Training on all pitchers does not improve performance, likely since left handed and right handed pitchers have very different throwing motions.

\begin{table}[]
    \centering
    \begin{tabular}{l|cccc}
        \toprule
        Arm & Acc & Prec & Rec & $F_1$  \\
        \midrule
        Lefty & .95 & .85 & .77 & .79 \\
        Righty & .94 & .81 & .74 & .74 \\
        All pitchers & .91 & .75 & .73 & .74 \\
         \bottomrule
    \end{tabular}
    \caption{Classification of pitcher injury trained on the 12 left handed or 8 right handed pitchers and evaluated on held-out data.}
    \label{tab:by-arm1}
\end{table}

\paragraph{Unseen pitchers} In Table \ref{tab:by-arm2} we report results for a model trained on 6 left handed (or 4 right handed) pitchers and tested on the other 6 left handed (or 4 right handed) pitchers not seen during training. For these experiments, the last 20 pitches thrown were considered `injured.' We find that when training with more pitcher data, the model generalizes better than when transferring from a single pitcher, but still performs quite poorly. Further, training on both left handed and right handed pitchers reduces performance. This suggests that models will be unable to predict injuries for pitchers they have not seen before, and left handed and right handed pitchers should be treated separately.

\begin{table}[]
    \centering
    \begin{tabular}{l|cccc}
        \toprule
        Arm & Acc & Prec & Rec & $F_1$  \\
        \midrule
        Lefty & .42 & .25 & .62 & .43 \\
        Righty & .38 & .22 & .54 & .38 \\
        All pitchers & .58 & .28 & .43 & .35 \\
        \bottomrule
    \end{tabular}
    \caption{Classification of pitcher injury trained on the 6 left handed or 4 right handed pitchers and evaluated on the other 6 left handed or 4 right handed pitcher.}
    \label{tab:by-arm2}
\end{table}

To determine if a model needs to see an specific pitcher injured before it can detect that pitchers injury, we train a model with `healthy' and `injured' pitches from 6 left handed pitchers (4 right handed), and only `healthy' pitches from the other 6 left handed (4 right handed) pitchers. We use half of the unseen pitchers `healthy' pitches as training data and all 20 unseen `injured' plus the other half of the unseen `healthy' pitches as testing data. Our results are shown in Table \ref{tab:by-arm3}, confirming that training in this method generalizes to the unseen pitcher injuries, nearly matching the performance of the models trained on all the pitchers (Table \ref{tab:by-arm1}). This suggests that the models can predict pitcher injuries even without seeing a specific pitcher with an injury.

\begin{table}[]
    \centering
    \begin{tabular}{l|cccc}
        \toprule
        Arm & Acc & Prec & Rec & $F_1$  \\
        \midrule
        Lefty & .67 & .62 & .65 & .63 \\
        Righty & .71 & .65 & .68 & .67 \\
        All pitchers & .82 & .69 & .72 & .71 \\
         \bottomrule
    \end{tabular}
    \caption{Classification performance of our model when trained using `healthy' + `injured' pitch data from half of the pitchers  and some `healthy' pitches from the other half. The model was applied to unseen pitches from the other half of the pitchers to measure the performance. This confirms that the model can generalize to \textbf{unseen pitcher injuries} using only `healthy' examples.}
    \label{tab:by-arm3}
\end{table}

\paragraph{Lefty vs Righty Models} To further test how well the model generalizes, we evaluate the model trained on left handed pitchers on the right handed pitchers, and similarly the right handed model on left handed pitchers.  We also try horizontally flipping the input images, effectively making a left handed pitcher appear as a right handed pitcher (and vice versa). Our results, shown in Table \ref{tab:by-arm4}, show that the learned models do not generalize to pitches throwing with the other arm, but by flipping the image, the models generalize significantly better, giving comparable performance to unseen pitchers (Table \ref{tab:by-arm2}). By additionally including flipped `healthy' pitches of the unseen pitchers, we can further improve performance. This suggests that flipping an image is sufficient to match the learned motion information of an injured pitcher throwing with the other arm.

\begin{table}
\small
    \centering
    \begin{tabular}{l|cccc}
        \toprule
        Arm & Acc & Prec & Rec & $F_1$  \\
        \midrule
        Left-to-Right & .22 & .05 & .02 & .03 \\
        Right-to-Left & .14 & .07 & .05 & .05 \\
        Left-to-Right + Flip & .27 & .35 & .42 & .38 \\
        Right-to-Left + Flip & .34 & .38 & .48 & .44 \\
        Left-to-Right + Flip + `Healthy' & .57 & .54 & .57 & .56 \\
        Right-to-Left + Flip + `Healthy' & .62 & .56 & .55 & .56 \\
         \bottomrule
    \end{tabular}
    \caption{Classification of the left-handed model tested on right handed pitchers and the right handed model tested on left handed pitchers. We also test horizontal flipping the images, which makes a left handed pitcher appear as a right handed pitcher (and vice versa). }
    \label{tab:by-arm4}
\end{table}

\begin{table*}
    \centering
    \begin{tabular}{l|cccc|cccc}
        \toprule
         & \multicolumn{4}{|c|}{Lefty}  & \multicolumn{4}{|c}{Righty}\\
        Injury & Acc & Prec & Rec & $F_1$  & Acc & Prec & Rec & $F_1$ \\
        \midrule
        Back Strain & .95 & .67 & .74 & .71 & .95 & .71 & .76 & .73 \\
        Finger Blister & .64 & .06 & .02 & .05 & .64 & .03 & .01 & .02 \\
        Shoulder Strain & .94 & .82 & .89 & .85 & .96 & .95 & .92 & .94 \\
        UCL Tear & .92 & .74 & .72 & .74 & .94 & .71 & .67 & .69 \\
        Intercostal Strain & .94 & .84 & .87 & .86 & .92 & .82 & .84 & .83 \\
        Sternoclavicular & .92 & .64 & .68 & .65 & .93 & .65 & .67 & .66 \\
        Rotator Cuff & .86 & .58 & .61 & .59 & .86 & .57 & .60 & .59 \\
        Hamstring Strain & .98 & .89 & .92 & .91 & .99 & .93 & .95 & .94 \\
        Groin Strain & .93 & .85 & .83 & .84 & .92 & .85 & .84 & .86 \\
       \bottomrule
    \end{tabular}
    \caption{Evaluation of the model for each injury. The model is able to detect some injuries well (e.g., hamstring strains) and others quite poorly (e.g., finger blisters).}
    \label{tab:by-inj}
\end{table*}


\begin{table*}
    \centering
    \begin{tabular}{l|ccccc}
        \toprule
         & $k=10$ & $k=20$ & $k=30$ & $k=50$ & $k=75$ \\
        \midrule
        Lefty & .68 & .63 & .65 & .48 & .47 \\
        Righty & .64 & .67 & .69 & .52 & .44 \\
       \bottomrule
    \end{tabular}
    \caption{$F_1$ score for models trained to predict the last $k$ pitches as `injured.' The model is able to produce most accurate predictions where the last 10 to 30 pitchers are labeled as `injured' but performs quite poorly for greater than 50 pitches.}
    \label{tab:how-early}
\end{table*}

\subsection{Analysis of Injury Type}
We can further analyze the models performance on specific injuries. The 10 injuries in our dataset are: back strain, arm strain, finger blister, shoulder strain, UCL tear, intercoastal strain, sternoclavicular joint, rotator cuff, hamstring strain, and groin strain. For this experiment, we train a separate model for left-handed and right-handed pitchers, then compare the evaluation metrics for each injury for each throwing arm. We use half the pitchers for training data plus half the `healthy' pitches from the other pitchers. We evaluate on the unseen `injured' pitches and other half of the unseen `healthy' pitches.

In Table \ref{tab:by-inj}, we show our results. Our model performs quite well for most injuries, especially hamstring and back injuries. These likely lead to the most noticeable changes in a pitchers motion, allowing the model to more easily determine if a pitcher is hurt. For some injuries, like finger blisters, our model performs quite poorly in detecting. Pitchers likely do not significantly change their motion due to a finger blister, as only the finger is affected.

\subsection{How early can an injury be detected?}
Following the best setting, we use half the pitchers plus half of the `healthy' pitches of the remaining pitchers as training data and evaluate on the remaining data (i.e., the setting used for Table \ref{tab:by-arm3}). We vary $k$, the number of pitches thrown before being placed on the disabled list to determine how early before injury the model can detect an injury. In Table \ref{tab:how-early}, we show our results. The models performs best when given 10-30 `injured' samples, and produces poor results when the last 50 or more pitches are labeled as `injured.' This suggests that 10-30 samples are enough to train the model while still containing sufficiently different motion patterns related to an injury. When using the last 50 or more pitches, the injury has not yet significantly impacted the pitchers throwing motion.

\section{Evaluating the Bias in the Dataset}

To confirm that our model is not fitting to game-specific data and that such game-specific information is not present in our optical flow input, we train an independent CNN model to predict which game a given pitch is from. The results, shown in Table \ref{tab:game-prediction}, show that when given cropped optical flow as input, the model is unable to determine which game a pitch is from, but is able to when given RGB features. This confirms both that our cropped flow is a good input and that the model is not fitting to game specific data. 

\begin{table*}[]
    \centering
    \begin{tabular}{l|ccccc}
        \toprule
        Pitcher & Guess & RGB & Cropped RGB & Flow & Cropped Flow \\
        \midrule
        Boone Logan  & 0.44 & 0.97 & 0.95 & 0.76 & 0.47  \\
        Clayton Kershaw & 0.44 & 0.94 & 0.85 & 0.73 & 0.45 \\
        Adalberto Mejia & 0.54 & 0.98 & 0.78 & 0.55 & 0.55 \\
        Aroldis Chapman & 0.25 & 0.86 & 0.74 & 0.57 & 0.26 \\
         \bottomrule
    \end{tabular}
    \caption{Accuracy predicting which game a pitch is from using different input features. Having a lower value means that the input data has less bias, which is better for the injury detection. The model is able to accurately determine the game using RGB features. However, using the cropped flow, it provides nearly random guess performance, confirming that when using cropped flow, the model is not fitting to game-specific details. Note that random guessing varies depending on how many games (and pitches per-game) are in the dataset for a given pitcher. }
    \label{tab:game-prediction}
\end{table*}

We further analyze the model to confirm that our input does not suffer from temporal bias, by trying to predict the temporal ordering of pitches. Here, we give the model two pitches as input, and it must predict if the first pitch occurs before or after the second pitch. We only train this model on pitches from games where there was no injury to confirm that the model is fitting to injury related motions, and not some other temporal feature. The results are shown in Table \ref{tab:order-prediction} and we find that the model is unable to predict temporal ordering of pitches. This suggests that the model is fitting to actual injury related motion, and not some other temporal feature.

\begin{table}[]
    \centering
    \begin{tabular}{l|c}
        \toprule
        Pitcher & Accuracy  \\
        \midrule
        Boone Logan  & 0.49  \\
        Clayton Kershaw & 0.53 \\
        Adalberto Mejia & 0.54 \\
        Aroldis Chapman & 0.51 \\
        \midrule
        All Average & 0.51\\
         \bottomrule
    \end{tabular}
    \caption{Predicting if a pitch occurs before or after another pitch, random guess is 50\%. We used only pitches from games where no injury occurred to determine if the model was able to find any temporal relationship between the pitches. Ideally, this accuracy would be 0.5, meaning that the pitch ordering is random. We find the model is not able to predict the ordering of pitches, suggesting that it is fitting to actual different motions caused by injury and not unrelated temporal data.}
    \label{tab:order-prediction}
\end{table}

\section{Discussion and Conclusions}

\begin{figure}
    \centering
    \includegraphics[width=\linewidth]{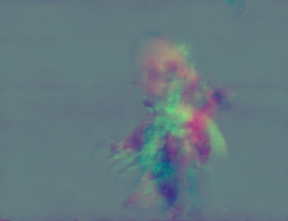}
    \caption{Visualization using the method from \cite{feichtenhofer2018have} to visualize an `injured' pitch. This does not give an interpretable image of why the decision was made, but shows that the model is capturing spatio-temporal pitching motions.}
    \label{fig:vis}
\end{figure}

We introduced the problem of detecting and predicting injuries in pitchers from only video data. However, there are many possible limitations and extensions to our work. While we showed that CNN can reliably detect and predicted injuries, due to the somewhat limited size of our dataset and scarcity of injury data in general, it is not clear exactly how well this will generalize to all pitchers, or pitchers at different levels of baseball (e.g., high school pitchers throw much more slowly than the professionals). While optical flow provides a reasonable input feature, it does lose some detail information which could be beneficial for injury detection. The use of higher resolution and higher frame-rate data could further improve performance. Further, since our method is based on CNNs, it is extremely difficult to determine why or how a decision is made. We applied the visualization method from Feichtenhofer et al. \cite{feichtenhofer2018have} to our model and data to try to interpret why a certain pitch was classified as an injury. However, this just provided a rough visualization over the pitchers throwing motion, providing no real insight into the decision. We show an example visualization in Fig. \ref{fig:vis}. It confirms the model is capturing spatio-temporal pitching motions, but does not explain why or how the model detects injuries. This is perhaps the largest limitation of our work (and CNN-based methods in general), as just a classification score is very limited information for the athletes and trainers.

As many injuries in pitchers are due to overuse, representing an injury as a sequence of pitches could be beneficial, rather than treating each pitch as an individual event. This would allow for models to detect changes in motion or form over time, leading to better predictions and possibly more interpretable decisions. However, training such sequential models would require far more injury data to learn from, as 10-20 samples would not be enough. The use of additional data, both `healthy' and `injured' would further improve performance. Determining the optimal inputs and designing of models specific to baseball/pitcher data could further help.

Finally, determining how early and injury would have to be detected/predicted to actually reduce recovery time remains unknown.

In conclusion, we proposed a new problem of detecting/predicting injuries in pitchers from only video data. We extensively evaluated the approach to determine how well it performs and generalizes for various pitchers, injuries, and how early reliable detection can be done.

{\small
\bibliographystyle{ieee}
\bibliography{egbib}
}

\end{document}